%% file: main.tex
\documentclass[10pt,twocolumn,letterpaper]{article}

\usepackage{cvpr}              
\usepackage[accsupp]{axessibility} 
\input{preamble}

\definecolor{cvprblue}{rgb}{0.21,0.49,0.74}
\usepackage[pagebackref,breaklinks,colorlinks,citecolor=cvprblue]{hyperref}
\usepackage{adjustbox}      
\usepackage{multirow}
\usepackage{gensymb}        
\def\paperID{9} 
\def\confName{VOCVALC}
\def\confYear{2024}

\title{\LARGE \bf
BAA\includegraphics[width=1em]{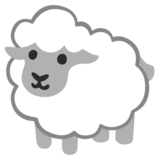}-NGP: Bundle-Adjusting Accelerated Neural Graphics Primitives}

\author{Sainan Liu\thanks{Equal contributions.}\\
Intel Labs\\
{\tt\small sainan.liu@intel.com}
\and
Shan Lin\footnotemark[1]\\
UCSD\\
{\tt\small shl102@ucsd.edu}
\and
Jingpei Lu\\
UCSD\\
{\tt\small jil360@ucsd.edu}
\and
Alexey Supikov\\
Intel Labs\\
{\tt\small alexei.soupikov@intel.com}
\and
Michael Yip\\
UCSD\\
{\tt\small yip@ucsd.edu}
}

\begin{document}
\maketitle
\input{sec/0_abstract}    
\input{sec/1_intro}
\input{sec/2_methodology}
\input{sec/3_experiments}
\input{sec/4_conclusions}
\input{sec/5_acknowledgement}

{
    \small
    \bibliographystyle{ieeenat_fullname}
    \bibliography{main}
}


\end{document}

%% file: preamble.tex
\usepackage[dvipsnames]{xcolor}


%% file: sec/0_abstract.tex
\begin{abstract}
Implicit neural representations have become pivotal in robotic perception, enabling robots to comprehend 3D environments from 2D images. Given a set of camera poses and associated images, the models can be trained to synthesize novel, unseen views. To successfully navigate and interact in dynamic settings, robots require the understanding of their spatial surroundings driven by unassisted reconstruction of 3D scenes and camera poses from real-time video footage. Existing approaches like COLMAP and bundle-adjusting neural radiance field methods take hours to days to process due to the high computational demands of feature matching, dense point sampling, and training of a multi-layer perceptron structure with a large number of parameters. To address these challenges, we propose a framework called bundle-adjusting accelerated neural graphics primitives (BAA-NGP) which leverages accelerated sampling and hash encoding to expedite automatic pose refinement/estimation and 3D scene reconstruction. Experimental results demonstrate 10 to 20 $\times$ speed improvement compared to other bundle-adjusting neural radiance field methods without sacrificing the quality of pose estimation. The github repository can be found here \href{https://github.com/IntelLabs/baa-ngp}{https://github.com/IntelLabs/baa-ngp}.
\end{abstract}

%% file: sec/1_intro.tex
\section{Introduction}
\label{sec:intro}

As robotics continues to push the boundaries of automation and intelligence, the incorporation of advanced computer vision techniques has become indispensable. The effectiveness of robots in the real world relies on their ability to localize, perceive, and model the 3D world from cameras with high accuracy and high resolution. This perception challenge has been ever-present in robotics and automation research for decades. In recent years, implicit neural representation (INR) \cite{curless1996volumetric} has shown great success in capturing highly detailed and accurate 3D reconstructions of objects and scenes with neural networks. Given a set of known camera intrinsic and extrinsic parameters and a rough estimation of the boundary of the scene, an implicit 3D scene representation can be learned by sampling 3D points along camera rays and performing supervised learning using ground-truth color images \cite{mildenhall2020nerf}. Such INR can accurately model scene geometry, radiance, and material properties providing key data for robotics tasks like localization, navigation, and object manipulation ~\cite{maggio2023loc, zhu2023latitude, tang2023rgb, dai2023graspnerf, adamkiewicz2022vision}.

Using INR in robotics poses unique challenges. The need for accurate camera poses is one of them. While popular methods like COLMAP~\cite{schoenberger2016sfm, schoenberger2016mvs} can estimate the camera poses from input images, they often drop frames that lack distinctive features and lead to suboptimal results. Some approaches propose to simultaneously optimize pose estimations and learn the radiance field~\cite{lin2021barf, chng2022garf}. Yet, their computational cost, leading to hours of training on just 100 images of size $400\times400$, makes them unsuitable for real-time applications. 

Recent works focused on making INR learning faster. For example, instant neural graphics primitives (iNGP) \cite{mueller2022instant} uses occupancy grid and hash encodings to converge significantly faster ($<5$ seconds) than other techniques with ground truth poses. However, this approach was not shown to work without or with poorly estimated camera poses, making it less applicable in many real-world, roaming camera scenarios.

\begin{figure}[t]
  \centering
  \includegraphics[width=0.85\linewidth]{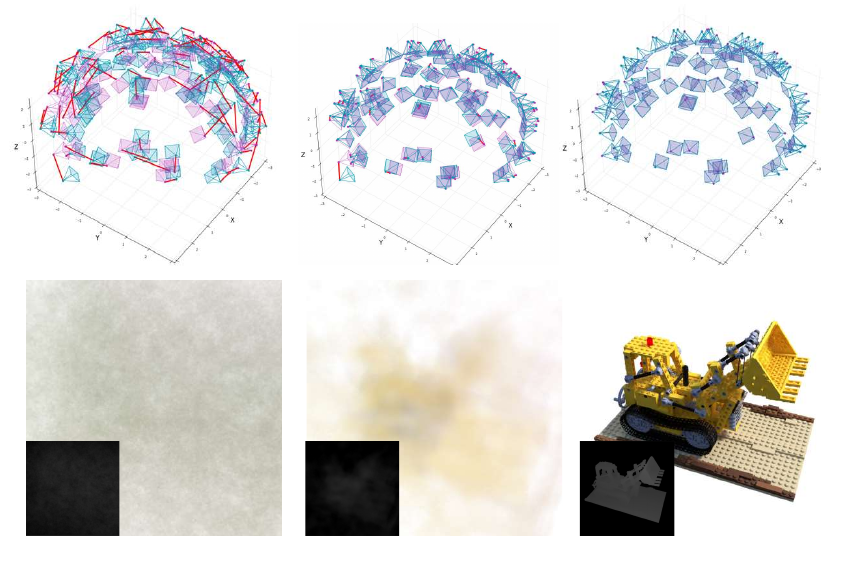}
  \scriptsize
  \begin{tabular}{ccc}
~~~~  Initial poses ~~~ & ~~~ Intermediate poses & ~~~ Optimized poses \\
~~~~  0 iteration ~~~& ~~~ 2k iterations & ~~~ 40k iterations \\

  \end{tabular}
  \caption{BAA-NGP is a neural implicit representation that captures 3D scenes from 2D images with unknown camera poses. It learns the 3D scene together with the camera poses within minutes of training, whereas previous methods would have taken hours.}
 \vspace{-2em}
  \label{fig:teaser}
\end{figure}

In this paper, we propose a novel approach called Bundle-Adjusting Accelerated Graphics Primitives (BAA-NGP) that estimates camera poses and optimizes the radiance field simultaneously using a computationally optimized approach that results in a 10 to 20 times speedup Figure~\ref{fig:teaser}. This fundamentally addresses the challenge of accelerated learning of INR models in unstructured settings without tracked cameras, making the method widely applicable to many real-world applications. BAA-NGP works by combining pose estimation with fast occupancy sampling and multiresolution hash encoding through a new curriculum learning strategy. We evaluate our proposed method on several benchmark datasets, including multi-view object-centric scenes \cite{mildenhall2020nerf} as well as frontal-camera video sequences of unbounded scenes \cite{mildenhall2019local}. We compare our results with state-of-the-art techniques 
and show that we achieve  
comparable or better performance while being significantly faster to learn. This gives BAA-NGP a range of applicability to a broad set of real-world scenarios and applications ranging from robotics and automation to virtual and augmented reality. 

\section{RELATED WORK}
Despite the relatively recent development of INRs, an expansive number of approaches have been proposed in the literature. Therefore we will focus on works related to solving for INRs with noisy/poor camera poses. 

\subsection{Camera Pose Estimation for 3D Perception}
Structure from motion (SfM) is a classical method often used for 3D structure reconstruction from 2D images or video sequences involving estimating camera poses for each image and simultaneously recovering 3D positions of the scene points. They leveraging feature detection and matching across frames. The camera pose estimation techniques from SfM have been used in the context of INR training. COLMAP~\cite{schoenberger2016sfm, schoenberger2016mvs} is a commonly used library for performing SfM, and many INR methods rely on this as part of their pipeline. However, COLMAP and other SfM techniques have high computational and memory requirements and rely heavily on an abundance of salient features in the scene, which may result in missing frames in scenes with limited or poor feature detection/saliency.

To avoid the limitations associated with using SfM for camera pose estimation, NeRF$--$~\cite{wang2021nerf--} and Self-Calibrating Neural Radiance Fields \cite{jeong2021self} learned camera pose estimates along with the neural radiance fields. These methods assumed forward-facing views and involved two-stage procedures for estimating camera poses and updating the neural radiance fields during training. Bundle-adjusting neural radiance field (BARF)~\cite{lin2021barf} first proposed a direct extrinsics estimation while learning the implicit 3D scene with NeRF. A coarse-to-fine feature weighting schedule for positional encoding features was used and was critical for smoothing the signals for pose updates. G-Nerf~\cite{meng2021gnerf} used GANs to improve generalization to large baseline captures. Gaussian-Activated Radiance Fields~\cite{chng2022garf} then proposed that without positional encoding, updating the activation function of the multi-layer perceptron can further improve the pose estimation and the quality of the results.\cite{chen2023local} proposed to further constrain pose estimation with a local deformation field, which improved the convergence quality for gradual camera movement. However, these methods still require an extensive amount of time to train (in the low tens of hours). 

\subsection{Approaches to Reduce INR Learning Time}
To overcome the long training time of INR learning, recently, several methods have been proposed to accelerate the process from the aspect of sampling, encoding, and better hardware integration of the multilayer perceptron, including but not limited to \cite{piala2021terminerf,fang2021neusample,yu2021plenoxels,wu2022diver,sun2022direct}. However, these methods assume known camera poses. Methods that train on scenes without camera poses still rely on SfM, such as COLMAP, for pose estimation. \cite{yen2020inerf, lin2023pnerf} has utilized faster architecture for pose retrieval, but their task is to align a novel view with an existing implicit 3D object, whereas we are learning the 3D representation from scratch. \cite{Zhu22nice, sucar21imap} are real-time SLAM methods that utilize implicit neural rendering techniques, but they both require extra depth as part of the input. Instead, a recent paper \cite{Heo2023robust} presented a fast neural radiance field learning without a camera prior that avoided classical SfM methods. This work utilizes gradient smoothing and re-implemented iNGP in PyTorch in order to use multi-level learning rate scheduling to incorporate multi-resolution hash encoding with pose estimation. Unfortunately, the code is unavailable, so the performance cannot be cross-validated. 
\\
\\
We focus on improving the speed of the basic bundle-adjusting NeRF structure. Our work approaches the problem via a simple and novel coarse-to-fine feature re-weighting scheme which enables pose refinement capability for hash encoding with multi-resolution occupancy sampling, resulting in 10 $\times$ shorter training iterations than the concurrent work \cite{Heo2023robust} and 10 to 20 $\times$ speedup on two benchmark datasets comparing to BARF \cite{lin2021barf}. For unbounded scenes, we incorporate inverted sphere parameterization with hash encoding, which enables pose estimation for more general scenarios.

%% file: sec/2_methodology.tex
\section{Methodology}
\label{sec:methodology}

\subsection{Problem Formulation}
Our goal is to learn INR models from images taken without known camera poses or poorly estimated poses. The INR model is described in general terms by a function
\begin{equation}
F_\Theta(p): \{x,y,z\} \rightarrow c, \sigma
\end{equation}
where a 3D location in the scene $\{x,y,z\}$ maps to a 3-channel radiance color $c=\{r,g,b\}$ and density $\sigma\in[0,1]$. In order to reconstruct an image from a camera positioned and oriented as $p\in\mathbb{S}\mathbb{E}(3)$ in the scene, we apply a pixel-by-pixel ray marching procedure, where for each pixel location $[u,v]$, we integrate all the radiance values found along the ray going from the camera origin and passing through the pixel location $[u,v]$. Therefore, the output color of a pixel along the ray $R$ of a camera with pose $p$ can be defined as:
    \begin{equation}
        f(p, R) = \sum\nolimits_{i=1}^N \alpha_i T_i c_i
    \end{equation}
where $T_i  = exp(- \sum_{j=1}^{i-1} \sigma_j \delta_j)$ is the accumulated transmittance along the ray, and $c_i$ is the output color of sample $i$ from $F_\Theta(p)$, and $\alpha$ for each ray segment is calculated as 
    \begin{equation}
        \label{eq:alpha-composition}
        \alpha_i = 1 - exp(\sigma_i\delta_i)
    \end{equation}
where $\delta_i$ is the distance between sample point $i$ and $i+1$. $\sigma_i$ is the density output of sample point $i$. 

For training, given $M$ captured images $\{ \mathcal{I} _i\}_{i=1}^M$ of width $W$ and height $H$ taken from the same scene, we optimize the INR model, and simultaneously, as a byproduct, optimize the camera poses $\{p_i\}_{i=1}^M \in\mathbb{R}^6$ associated with each image.

\begin{figure}
  \centering
  \includegraphics[width=\linewidth]{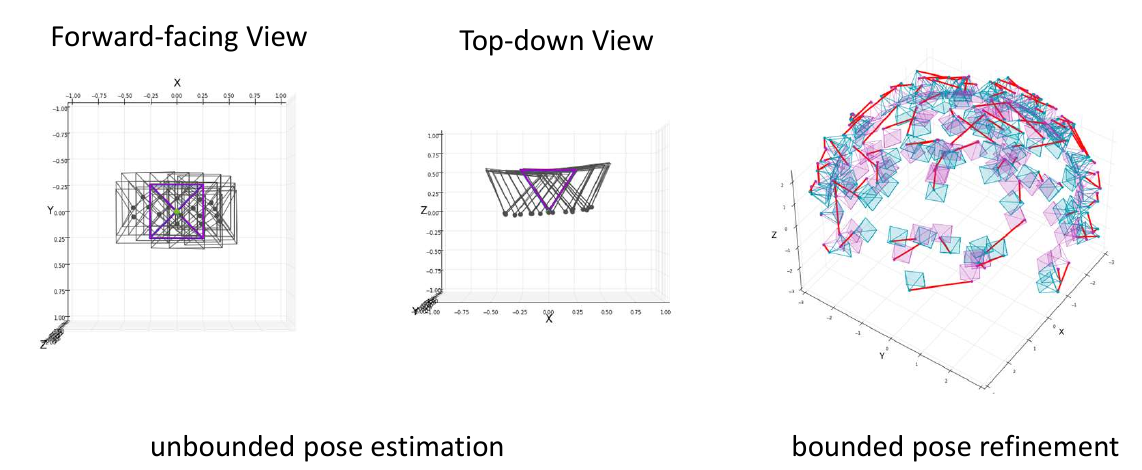}
  \vspace{-2em}
  \caption{INRs use posed images from multiple viewpoints to reconstruct the scene. In our problem, we assume that a sequence of images was taken from unknown viewpoints for unbounded scenes (left) and poorly estimated viewpoints for bounded scenes (right). Purple frames are initial camera poses, gray/blue frames are ground truth camera poses, and the red line indicates a translation error.}
 \label{fig:nerf_concept_figure}
  \vspace{-1em}
\end{figure}

Similar to \cite{lin2021barf} and \cite{chng2022garf}, we have no image order or sequence assumptions. We assume that camera intrinsics are known, but by definition, camera extrinsics are unknown. We also assume that the scene is static. Solving the problem involves minimizing the following loss function:
\begin{equation}
    \min_{p_1, ..., p_M, \Theta} \sum_{i=1}^M\sum_{j=1}^W\sum_{k=1}^H||\hat{\mathcal{I} }_{ijk}(p_i, \Theta) - \mathcal{I} _{ijk}||_2^2
    \label{eq:general_objective}
\end{equation}
The camera poses $\{p_1 ... p_M\}$ are optimization variables and are found jointly with the model parameters $\Theta$. Previous works involving camera pose estimation \cite{lin2021barf,chng2022garf} have considered the following two scenarios, each with different datasets and objectives (Figure~\ref{fig:nerf_concept_figure}):
\begin{itemize}
    \item \textit{Pose estimation (unbounded scene)}: Camera poses are reconstructed from video with frames captured in sequence relatively close to each other. This ``forward-facing scene'' data is often from cellphone cameras or mobile robots. 
    \item \textit{Pose refinement (bounded scene)}: camera pose corrections are found for available but noisy camera poses for multi-viewpoint images of a centered object.
\end{itemize}
Small variations in the model configuration are used to account for these two domain types. 

\subsection{Approach}
\subsubsection{Network Architecture} 
We start with an established model for learning implicit neural representations, i.e., neural radiance field, consisting of a neural network model for representing $F_\theta(p)$. One of the most successful models for this task is an MLP. Our final model involves the following stages for image reconstruction: inverted sphere parameterization, multi-resolution hash encoding, then an MLP decoder.

    \begin{figure}
    \centering
    \includegraphics[width=0.5\linewidth]{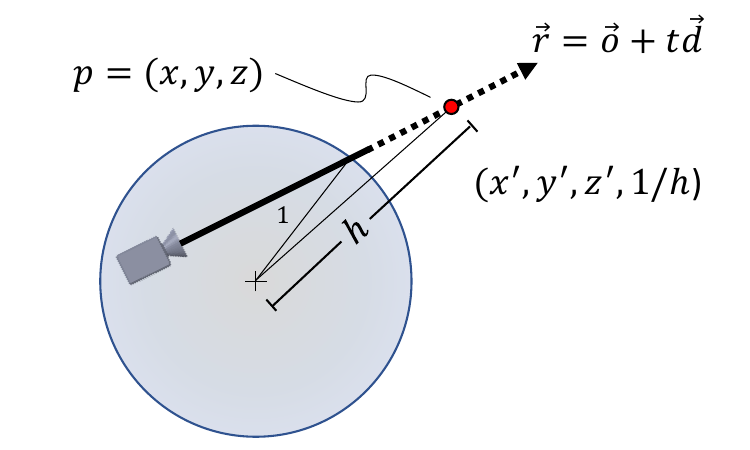}
    \vspace{-3mm}
    \caption{Reparameterization of 3D space via spherical contraction. Each point $p=(x,y,z)$  outside unit sphere becomes $(x',y',z',1/h)$, a quadruple that converts unbounded distances $h$ to bounded distances $1/h$.}
    \vspace{-1em}
    \end{figure}

\emph{\textbf{Parameterized camera poses to contracted 3D points}}:
Since hash grids and occupancy grids need finite bounding boxes known in advance, handling bounded and unbounded scenes requires different 3D space parameterizations. For bounded scenes with a given bounding box, we use affine transformations to bring the volume of interest into an axis-aligned bounding box $[0,1]^3$. For unbounded scenes, we apply an inverted sphere parameterization allowing compact mapping of close and far locations \cite{zhang2020nerf++}.
A point $(x, y, z)$ seen by a virtual camera with camera center $o\in\mathcal{R}^3$ along a ray $r=o+td$ with unit direction $d\in\mathcal{R}^3$, and with distance $t=[0,\infty]$, can be located arbitrarily far. 
To address this, a point within the unit sphere, centered at the origin, is unchanged. A point outside of the volume $h = \sqrt{x^2 + y^2 + z^2} > 1$ is reparameterized to quadruple $(x', y', z', 1/h)$, where $(x', y', z')=(x,y,z)/h$, and $x'^2 + y'^2 + z'^2 = 1$. Affine transformations are used to bring points within the volume into $[0,1]^3$ and reparameterized 4D representation into $[0,1]^4$.

\emph{\textbf{Multiresolution Hash Encoding}}:
    We use multi-resolution hash encoding \cite{mueller2022instant} to speed up the radiance field learning.
    It transforms every input 3D point into a higher dimension concatenated features learned from each resolution level. For each point, $x \in \mathbb{R}^d$, the method first finds the grid cell in which the point resides, then $d$-linearly interpolates the features at vertices of the grid cell to produce an F-dimensional feature vector for the point $x$. Features from multiple resolution levels are then concatenated into a single vector. Instead of explicitly storing vertex features in a regular grid of size $N$ along each dimension, which would result in $N^d$ features for a level, hash encoding fixes the total number of F-dimensional features per level to be T. Hence if a coarser level needs less than $T$ features, the mapping remains 1:1, but if a finer level requires more than $T$ features, a spatial hash function (eq.~\ref{eq:hash_function}) \cite{Teschner2003spatialhash} is then used to map them into $T$ sets of features.
    \begin{equation}
    h(x) = (\bigoplus_{i=1}^d x_i \pi_i)\ mod\ T
    \label{eq:hash_function}
    \end{equation}
    where $x$ denotes the vertex position in d dimension, $\bigoplus$ indicates the bitwise XOR, and $\pi_1 = 1, \pi_2 = 2654435761, \pi_3 = 805459861$ are selected by the original authors \cite{mueller2022instant}.

    By specifying the base resolution ($N_{min}$) and the maximum resolution ($N_{max}$) of the grid, for $L$ levels, the grid resolution in between is chosen by $N_l = \lfloor N_{min} \cdot b^l \rfloor$, where b is calculated based on Equation~\ref{eq:hash_grid}.
    \begin{equation}
        \label{eq:hash_grid}
        b = exp((ln N_{max} - ln N_{min})/(L - 1))
    \end{equation}
    Value along a dimension of a point with range $[0,1]$ can be mapped to a certain grid cell along that dimension within the $l$ layers via $\lfloor{x_l} \rfloor$ and $\lceil{x_l} \rceil$, where $x_l = x \cdot N_l$. The $d$-linear interpolation weight for that dimension can then be found, for example, via $w_l = x_l - \lfloor x_l \rfloor$ for that vertex corner.
    
\emph{\textbf{Multi-layer Perceptron (MLP)}}: 
    MLPs are commonly used for capturing neural radiance fields. We incorporate a fully fused implementation of MLPs\cite{mueller2022instant} with one layer for density learning and two layers for color learning, which takes the output of the density layer as well as the view directions. The MLPs takes in the encoded learnable features from multi-resolution hash encoding, concatenated with the direction encoded with spherical harmonics transform, and outputs color and density for each 3D query point.

\subsection{Training Strategies}

\subsubsection{Occupancy Grid Sampling}
For further acceleration, we adopted coarse multiscale grid sampling before passing points through the radiance field for gradient backpropagation. Our occupancy grid is defined around the origin with an axis-aligned bounding box. The density of the sampled points is first queried without gradient to check if they hit free space. Only points that reside in a coarse grid cell that do not return free space (above a certain threshold) will be passed through for backpropagation.

\subsubsection{Signal Coarse-to-Fine Smoothing}
A coarse-to-fine smoothing technique has been proposed \cite{lin2021barf} to shield the earlier stage of pose estimation from being affected by high-frequency information and slowly introduce such information at later stages for improved pose fine-tuning and image reconstruction quality. They use positional encoding and weigh the $k$-th frequency components of the inputs with $w_k$:
\begin{equation}
    w_k (\alpha) =
    \begin{cases}
    0  & \text{if}\ \alpha < k\\
    \frac{1 - cos((\alpha - k)\pi)}{2} & \text{if}\ 0\leq \alpha - k < 1\\
    1 & \text{if}\ \alpha -k \geq 1 \\
    \end{cases} \\
\end{equation}
where $\alpha_i = \frac{s_i - r_s}{r_e - r_s}$. $r_s$ is the coarse-to-fine starting point as the percentage of the training progress, $r_e$ is the coarse-to-fine ending point as the percentage of the training progress, and $s$ is the percentage of training progress so far at step $i$. This method assumes the original coordinates are concatenated with the positional encoding features, allowing initial pose estimation with just 3D points as inputs. 
    
We propose a novel coarse-to-fine strategy better suited for learnable hash encoding features. Since hash grid features, accompanying the very small MLP, carry information about density and radiance distributions in the volume, they are not equivalent to scene-agnostic frequency encoding of the 3D coordinates. Using zero weights of the cosine window to nullify the features at finer levels results in MLP getting stuck at the early local minima and failing to learn hash grid features efficiently. Instead, starting with the coarsest level enabled, we prime the MLP by using coarse-level features in lieu of windowed-out fine-level features. We gradually replace such coarse-level estimates with the actually learned fine-level values as the cosine window expands. Thus, the features are weighted with
\begin{equation}
    \gamma_k(x;\alpha) = w_k(\alpha) \cdot d_k + ( 1 - w_k(\alpha) )\cdot d_{\alpha}
\end{equation}

where $d_\alpha$ indicates the set of features that has the highest grid level with a nonzero weight.

%% file: sec/3_experiments.tex
\section{Experiments}
\begin{table*}[ht!]
\caption{Quantitative evaluation of pose refinement for the blender synthetic dataset. BAA-NGP is on par with BARF\cite{lin2021barf} in terms of camera pose estimation, with 10x faster training time, and with better visual synthesis quality overall.}
  \label{tab:blender}
  \centering
  \tiny
  \begin{adjustbox}{width=\linewidth,center}
  \begin{tabular}{l||cc|cc||cc|cc|cc|cc|cc}
    \toprule
    \multirow{3}{*}{Scene}&\multicolumn{4}{c}{Camera pose registration}&\multicolumn{8}{c}{Visual synthesis quality}&\multicolumn{2}{c}{Training Time}\\
    & \multicolumn{2}{c}{Rotation(\degree) $\downarrow$} & \multicolumn{2}{c}{Translation $\downarrow$} & \multicolumn{2}{c}{PSNR $\uparrow$} &\multicolumn{2}{c}{SSIM $\uparrow$} &\multicolumn{2}{c}{MS-SSIM $\uparrow$}&\multicolumn{2}{c}{LPIPS $\downarrow$} &\multicolumn{2}{c}{hh:mm:ss}\\
    \cmidrule(r){2-15}
    &BARF&ours&BARF&ours&BARF&ours&BARF&ours&BARF&ours&BARF&ours&BARF&ours\\
    \midrule
    Chair & 0.093&\textbf{0.093} & \textbf{0.405}& 0.658 &31.15 & \textbf{34.36} & 0.952 & \textbf{0.985} & 0.990 & \textbf{0.996} & 0.074& \textbf{0.026} &04:26:30&\textbf{00:30:09}\\
    Drums & 0.046& \textbf{0.029} & 0.202& \textbf{0.134} & 23.91& \textbf{25.03} & 0.895& \textbf{0.934} & 0.954 & \textbf{0.971} & 0.147& \textbf{0.072}&04:28:06 &\textbf{00:26:10}\\
    Ficus & 0.080& \textbf{0.032} & 0.464& \textbf{0.161} & 26.26& \textbf{30.27} & 0.930& \textbf{0.979} & 0.975 & \textbf{0.991} & 0.109& \textbf{0.026}&04:21:56& \textbf{00:25:34}\\
    Hotdog & 0.229& \textbf{0.088} & 1.165& \textbf{0.529} & 34.59& \textbf{37.00} & 0.969& \textbf{0.982} & 0.992 & \textbf{0.994} & 0.059& \textbf{0.029}&04:24:52&\textbf{00:26:38}\\
    Lego & 0.081& \textbf{0.040} & 0.330& \textbf{0.144} & 28.31& \textbf{32.20} & 0.924& \textbf{0.975} & 0.981 & \textbf{0.993} & 0.106& \textbf{0.025}&04:24:21&\textbf{00:26:12} \\
    Materials &\textbf{0.837}& 1.021 & \textbf{2.703} & 4.944 &\textbf{27.85}& 27.16 & 0.934 & \textbf{0.943} & \textbf{0.984} & 0.983 & 0.107 & \textbf{0.077} &04:23:13&\textbf{00:23:46}\\
    Mic & 0.065& \textbf{0.046} &0.277 & \textbf{0.260} & 31.00 & \textbf{34.28} & 0.966 & \textbf{0.987} & 0.992 & \textbf{0.995} & 0.065& \textbf{0.018}&04:22:25& \textbf{00:32:46} \\
    Ship &0.086 & \textbf{0.061} & 0.341& \textbf{0.318} & 27.49& \textbf{29.71} & 0.841& \textbf{0.864} & 0.938 & \textbf{0.939} & 0.196& \textbf{0.123}&04:25:59& \textbf{00:23:31}\\
    \midrule
    Mean &0.190&\textbf{0.176}&\textbf{0.736}&0.894&28.82&\textbf{32.50}&0.926&\textbf{0.953}&0.976&\textbf{0.983}&0.108&\textbf{0.050}&04:24:40&\textbf{00:26:51}\\
    \bottomrule
  \end{tabular}
  \end{adjustbox}
\end{table*}
\subsection{Dataset}

We benchmarked BAA-NGP on the LLFF dataset for frontal-camera in-the-wild video sequences and the blender synthetic dataset for pose refinement. The following describes the dataset and the associated model architectures:

\emph{\textbf{LLFF}}
This video sequence dataset has images from continuous camera poses that concentrate the views at one side of the scene. The challenge of this dataset is that not many frames of each video sequence are available, and all cameras are posed on one side of the scene, giving limited coverage from multiple viewpoints. For training and evaluation, all images were resized to $640\times 480$.

For the network training setup, we used occupancy sampling, inverted sphere reparameterization for input points, and two sets of multiresolution hash encoding for in-sphere and outside spaces in addition to the vanilla MLP setup. Similar to \cite{lin2021barf}, all cameras are initialized with the identity transformation. We use the Adam optimizer and train the models for each scene with 20K iterations. The number of randomly sampled pixel rays is initialized as 1024 and dynamically adjusted throughout the training to keep the total number of samples taken along the rays consistent. The learning rate of the network is initialized as $1 \times 10^{-4}$ and linearly increased to $1 \times 10^{-2}$ for the first 100 iterations, and then step decayed by a factor of 0.33 at 10000, 15000, and 18000 iterations. Following a similar pattern, the learning rate of the camera poses is initialized as $3 \times 10^{-4}$ and linearly increased to $3 \times 10^{-3}$ for the first 100 iterations, and then step decayed by a factor of 0.33 at 10000, 15000, and 18000 iterations.

\emph{\textbf{Blender synthetic}}
The Blender synthetic dataset is a multi-view image dataset that includes imperfect camera pose estimations, for camera pose refinement and scene reconstruction. This dataset comprises 100 training and 200 testing images, which we resized to $400 \times 400$ for training and evaluation. The images are captured from the upper hemisphere of the central object and are rendered against a white background.

In terms of network training setup, we incorporate occupancy sampling, multiresolution hash encoding ($N_{min} = 14$, $N_{max} = 4069$, $L = 16$), and signal coarse-to-fine smoothing ($r_s = 0.1$, $r_e = 0.5$), with the vanilla MLP setup. The camera poses are perturbed by adding noise $\mathcal{N }(0, 0.15\mathbf{I})$ to the ground-truth poses, similar to the setup in Lin et al. \cite{lin2021barf}.
We trained the model for each scene over 40K iterations. We used the Adam optimizer and applied an exponentially decaying learning rate schedule, starting from $1 \times 10^{-2}$ and decaying to $1 \times 10^{-4}$ for the network, and $1 \times 10^{-3}$ decaying to $1 \times 10^{-5}$ for the camera poses.

\subsection{Metrics}
We benchmarked our experiments for both novel view image reconstruction quality and pose estimation accuracy. For novel view synthesis, peak signal-to-noise ratio (PSNR), structural similarity index (SSIM) \cite{zhang2018perceptual}, multi-scale structural similarity index (MS-SSIM). Learned perceptual image patch similarity (LPIPS) \cite{zhang2018perceptual} are used to evaluate the predicted image quality against the ground truth novel views. For pose estimation, we evaluate the camera pose rotation and translation differences between the predicted and the ground truth camera set after using Procrustes analysis for alignment.

\subsection{Results}
We build our architecture on top of nerfacc\cite{li2023nerfacc},
and use hash encoding from tiny-cuda-nn \cite{tinycudann}. An NVIDIA RTX3090 GPU is used for training and evaluation. 
\begin{table*}[!htp]
\caption{Quantitative evaluation of pose estimation for the LLFF dataset. We show that BAA-NGP is comparable to BARF with much less training time. iNGP\cite{mueller2022instant} results are included for reference.}
  \label{tab:llff}
  \centering
  \begin{adjustbox}{width=\linewidth,center}
  \begin{tabular}{l||cc|cc||ccc|ccc|ccc|ccc|ccc}
    \toprule
    \multirow{3}{*}{Scene}&\multicolumn{4}{c}{Camera pose registration}&\multicolumn{12}{c}{Visual synthesis quality}&\multicolumn{3}{c}{Training Time}\\
    & \multicolumn{2}{c}{Rotation(\degree) $\downarrow$} & \multicolumn{2}{c}{Translation $\downarrow$} & \multicolumn{3}{c}{PSNR $\uparrow$} &\multicolumn{3}{c}{SSIM $\uparrow$} &\multicolumn{3}{c}{MS-SSIM $\uparrow$}&\multicolumn{3}{c}{LPIPS $\downarrow$} &&&\\
    \cmidrule(r){2-20}
    &\cite{lin2021barf}&ours&\cite{lin2021barf}&ours&\cite{lin2021barf}&ours&iNGP\cite{mueller2022instant}&\cite{lin2021barf}&ours&iNGP\cite{mueller2022instant}&\cite{lin2021barf}&ours&iNGP\cite{mueller2022instant}&\cite{lin2021barf}&ours&iNGP\cite{mueller2022instant}&\cite{lin2021barf}&ours&iNGP\cite{mueller2022instant}\\
    \midrule
    Fern & \textbf{0.163} & 3.978 & \textbf{0.188} & 1.733 & \textbf{23.96} & 19.02 & 25.42 & \textbf{0.709} & 0.504 & 0.822 & \textbf{0.916}& 0.717 & 0.948 & \textbf{0.390} & 0.480 & 0.182 &05:23:29&\textbf{0:10:45}&0:07:33\\
    Flower & \textbf{0.224} & 2.258 & \textbf{0.233} & 0.530 & 24.07 & \textbf{25.52} & 26.78 & 0.712 & \textbf{0.811} & 0.851 & 0.891 & \textbf{0.935} & 0.952 & 0.379 & \textbf{0.157} & 0.139&05:24:55&\textbf{0:10:39} & 0:08:21\\
    Fortress & \textbf{0.460} & 0.786 & \textbf{0.352} & 0.733 & \textbf{28.86} & 28.59 & 28.54 & 0.816 & \textbf{0.825} & 0.882 & \textbf{0.950}& 0.946 & 0.967 & 0.266 & \textbf{0.203} & 0.113 &05:44:05&\textbf{0:11:16}&0:08:16\\
    Horns & \textbf{0.135} & 1.042 & \textbf{0.162} & 0.643 & \textbf{23.12} & 19.57 & 20.47 & \textbf{0.734} & 0.724 & 0.755 & \textbf{0.915}& 0.879 &  0.845 & 0.423 & \textbf{0.307} & 0.262 &05:22:02&\textbf{0:18:54}& 0:07:32\\
    Leaves & 1.274 & \textbf{1.249} & \textbf{0.253} & 0.342 & 18.67 & \textbf{20.14} & 20.80 & 0.529 & \textbf{0.687} & 0.738 & 0.831& \textbf{0.902} & 0.922 & 0.474& \textbf{0.264} & 0.243 & 06:47:43& \textbf{0:18:37} & 0:08:07\\
    Orchids & \textbf{0.629} & 6.110 & \textbf{0.409} & 2.956 & \textbf{19.37} & 12.28 & 19.60 & \textbf{0.570} & 0.143 & 0.682 & \textbf{0.853} & 0.243 & 0.888 & \textbf{0.423} & 0.608 & 0.220 & 05:14:03& \textbf{0:10:3 }& 0:07:26\\
    Rooms & \textbf{0.362} & 1.853 & \textbf{0.293} & 1.856 & \textbf{31.60} & 29.15 & 34.03 & \textbf{0.937} & 0.900 & 0.969 & \textbf{0.981} & 0.963 & 0.986 & \textbf{0.230} & 0.271 & 0.095&05:35:49&\textbf{0:20:23}&0:07:55\\
    T-rex & \textbf{1.030} & 1.749 & \textbf{0.641} & 1.005 & 22.32 & \textbf{23.41} & 25.18 & 0.771 & \textbf{0.861} & 0.892 & 0.927& \textbf{0.950} & 0.960 & 0.355 & \textbf{0.185} & 0.121 & 05:10:04 & \textbf{0:10:37} & 0:07:37\\
    \midrule
    Mean &\textbf{0.535}&2.378&\textbf{0.316}&1.225&\textbf{24.00}&22.21&25.10&\textbf{0.722}&0.682&0.824&\textbf{0.908}&0.817&0.934&0.369&\textbf{0.309}&0.172&05:35:16&\textbf{0:13:54}&0:08:53\\
    \bottomrule
  \end{tabular}
  \end{adjustbox}
\end{table*}

\begin{figure}[ht!]
  \centering
  \includegraphics[width=\linewidth]{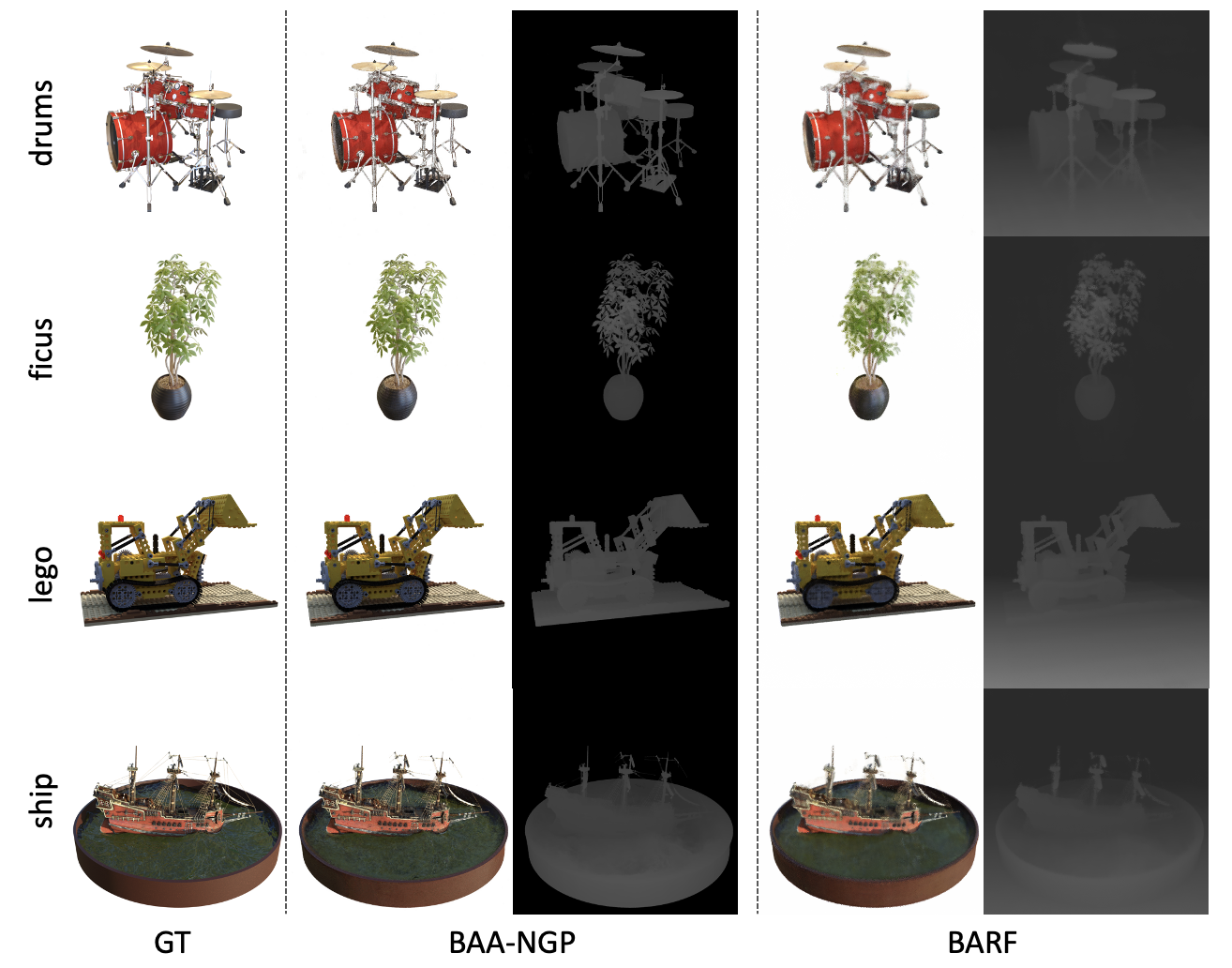}
  \caption{Qualitative analysis of BAA-NGP on the blender synthetic dataset. BAA-NGP produces better quality in image synthesis with cleaner backgrounds and finer details than BARF with 10 $\times$ less time.
  }
 \label{fig:blender}
 \vspace{-1em}
\end{figure}

\subsubsection{Unposed Frontal Camera Video Sequences}
\begin{figure*}[ht!]
  \centering
  \includegraphics[width=\linewidth]{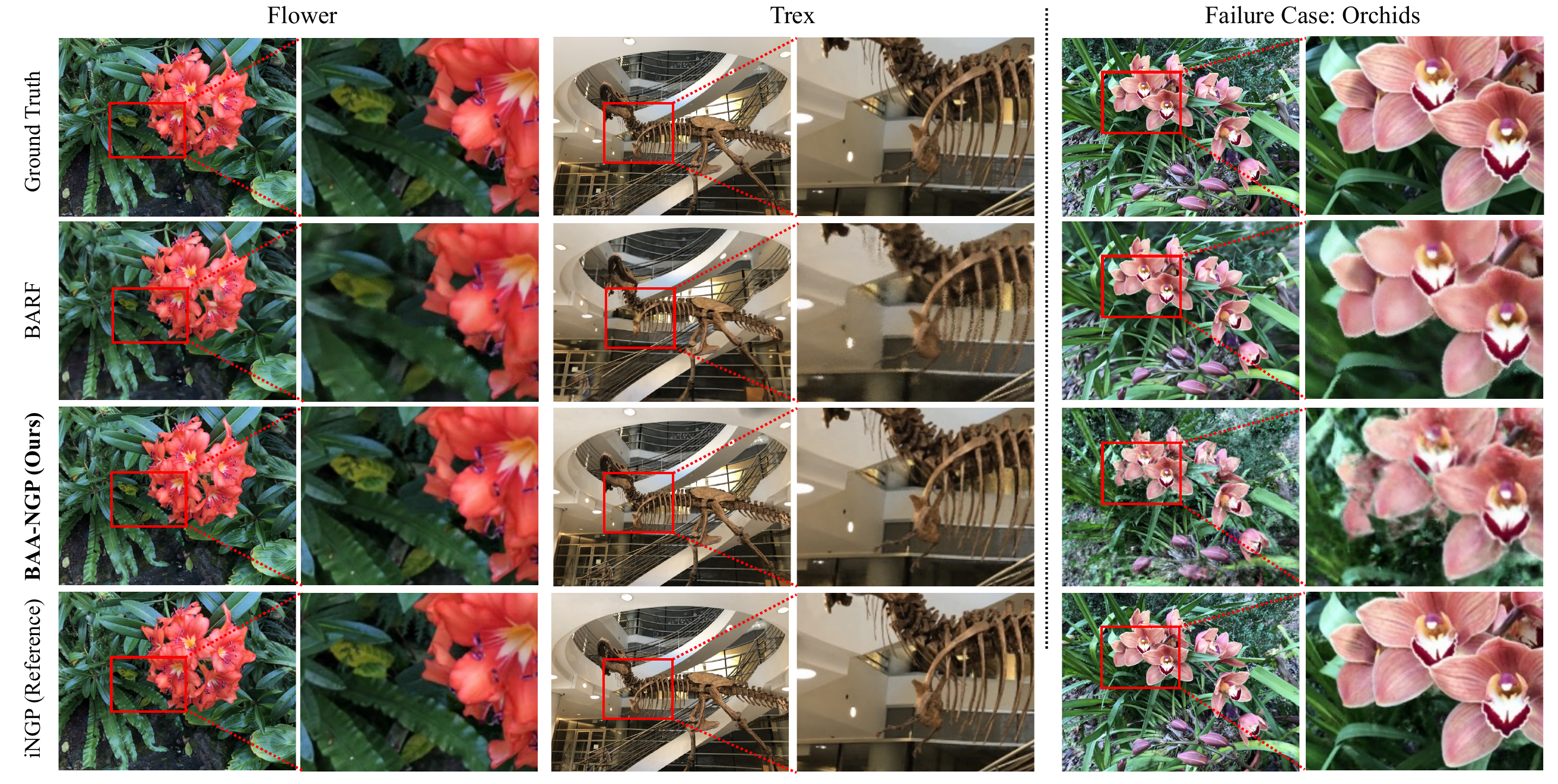}
 \vspace{-2em}
 \caption{Qualitative analysis of BAA-NGP on the LLFF dataset. We show that our results are on par with BARF's results but converge 20 $\times$ faster.}
 \label{fig:llff}
\end{figure*}

Figure~\ref{fig:llff} shows sample results of BAA-NGP on the frontal camera video sequences from LLFF, demonstrating qualitatively the high amount of detail in the reconstructed images. The last column of Figure~\ref{fig:llff} shows the only case (orchids) where our method fails to converge. It could be partially explained by the suboptimal performance of iNGP in this data, as shown in Table~\ref{tab:llff}. Quantitatively, Table~\ref{tab:llff} shows our benchmarking data against the state-of-the-art algorithms. We show that BAA-NGP can achieve more than $20 \times $ speedup with similar quality as BARF, whereas our metrics are better or on par for almost all metrics except rotation. Qualitatively, our method captures more details in the reconstructed images.

\subsubsection{Imperfect Poses Refinement for Multi-View Synthetic Images}
Figure~\ref{fig:blender} shows the results of imperfect pose refinement for multi-view synthetic images. Qualitatively, BAA-NGP produces better-quality image synthesis with cleaner backgrounds and finer details. For quantitative analysis, as shown in Table~\ref{tab:blender}, the results demonstrate that BAA-NPG not only yields similar pose estimation results and improved image synthesis quality but also achieves a more than 10-fold speedup. One observation from this result is that the Materials scenes in the dataset were more challenging, likely due to the limitations of hash grid feature encoding. The authors of iNGP reported similar findings when using ground truth camera poses, attributing this issue to the dataset's high complexity and view-dependent reflections \cite{mueller2022instant}. 

\begin{figure}[htp!]
  \centering
    \includegraphics[height=1.5in]{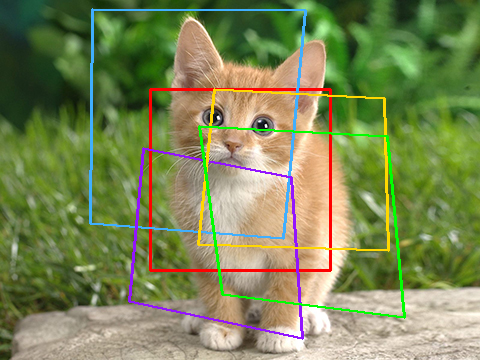}
    \includegraphics[height=1.5in]{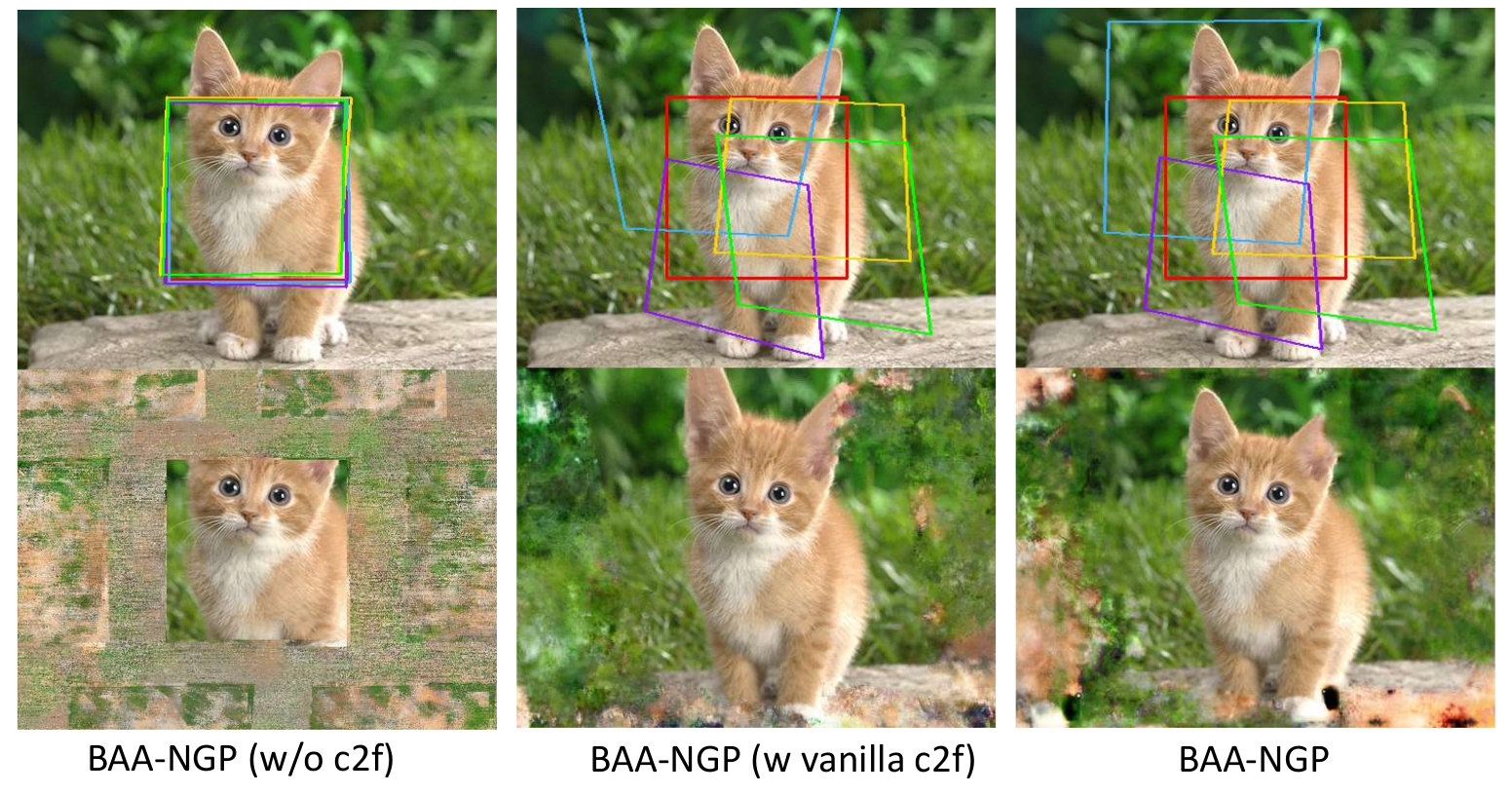}
  \caption{Homography recovery on a cat image. (L) Different colors indicate five different ground truth warped patches. (R) Results of different coarse-to-fine weighting procedures. Our coarse-to-fine weighting scheme outperforms the vanilla coarse-to-fine (c2f) weighting scheme qualitatively for both 2D homography recovery.}
 \label{fig:c2f}
 \vspace{-1em}
\end{figure}
\subsubsection{Homography Recovery}
We also conducted a homography recovery experiment as an analogy to 3D cases to evaluate the effectiveness of homography matrix estimation using multi-resolution hash encoding architecture combined with the MLP. We can consider equation~\ref{eq:general_objective} as a general case for homography recovery if we define $p_i$ as the parameterized warp transformation. We used the same setup as bundle-adjusting neural radiance field \cite{lin2021barf}, where five warped patches are used for training as is shown in Figure~\ref{fig:c2f} (L). 2D multi-resolution hash encoding and fully fused MLPs were used as the main network architecture, and coarse-to-fine feature weighting was used for the best performance. For this experiment, $N_{min} = 3, L = 18$, coarse-to-fine $r_s = 0.1, r_e = 0.5$, learning rate for features and parameters was $1.e-2$ and for warp parameters was $3.e-3$. Adam optimizer was used for both learning processes, and training was conducted over 5000 epochs.

We show in Figure~\ref{fig:c2f} (R) that without signal coarse-to-fine smoothing, the learned features of hash encoding converge without learning any warping functions, which end up learning the most overlapped area. Our coarse-to-fine implementation can achieve better pose estimation comparing to vanilla coarse-to-fine masking, which can be trapped in local minimal.

%% file: sec/4_conclusions.tex

\section{Conclusions}
In conclusion, we presented a method for learning INRs of scenes with unknown or poorly known camera poses while achieving learning rates in minutes. BAA-NGP, therefore, is a solution for addressing accelerated learning of INR models in unstructured settings. Our evaluations over several benchmark datasets, including multi-view object-centric scenes as well as frontal-camera video sequences of unbounded scenes, show that it is either comparable or outperforms state-of-the-art techniques such as BARF and COLMAP-based methods where camera poses are not known or are known imprecisely, and where training would otherwise take hours. Thus, this approach opens learning INRs for a broad set of real-world scenarios and applications ranging from virtual and augmented reality to robotics and automation, where time and unstructured image capture are vital. 

One observation is that when our rendering baseline iNGP leads to suboptimal results, the corresponding results of BAA-NGP will also be influenced. In the future, we plan to improve this technique and integrate BAA-NGP into robotic systems to provide fast and accurate environment modeling for real-time applications.


%% file: sec/5_acknowledgement.tex

\section*{Acknowledgment}
We express our heartfelt appreciation to Subarna Tripathi, Ilke Demir, Saurav Sahay, along with numerous other colleagues from Intel Labs as well as Shreya Saha from UCSD. 
Their insightful contributions, stimulating discussions, and generous provision of computing resources have been invaluable to our project.  

\clearpage
\setcounter{page}{1}

%% file: main.bbl
\begin{thebibliography}{32}
\providecommand{\natexlab}[1]{#1}
\providecommand{\url}[1]{\texttt{#1}}
\expandafter\ifx\csname urlstyle\endcsname\relax
  \providecommand{\doi}[1]{doi: #1}\else
  \providecommand{\doi}{doi: \begingroup \urlstyle{rm}\Url}\fi

\bibitem[Adamkiewicz et~al.(2022)Adamkiewicz, Chen, Caccavale, Gardner, Culbertson, Bohg, and Schwager]{adamkiewicz2022vision}
Michal Adamkiewicz, Timothy Chen, Adam Caccavale, Rachel Gardner, Preston Culbertson, Jeannette Bohg, and Mac Schwager.
\newblock Vision-only robot navigation in a neural radiance world.
\newblock \emph{IEEE Robot. Autom. Lett.}, 7\penalty0 (2):\penalty0 4606--4613, 2022.

\bibitem[Chen et~al.(2023)Chen, Chen, Wang, Zhang, Guo, Shan, and Wang]{chen2023local}
Yue Chen, Xingyu Chen, Xuan Wang, Qi Zhang, Yu Guo, Ying Shan, and Fei Wang.
\newblock Local-to-global registration for bundle-adjusting neural radiance fields.
\newblock In \emph{CVPR}, pages 8264--8273, 2023.

\bibitem[Chng et~al.(2022)Chng, Ramasinghe, Sherrah, and Lucey]{chng2022garf}
Shin-Fang Chng, Sameera Ramasinghe, Jamie Sherrah, and Simon Lucey.
\newblock Garf: Gaussian activated radiance fields for high fidelity reconstruction and pose estimation.
\newblock \emph{ECCV}, 2022.

\bibitem[Curless and Levoy(1996)]{curless1996volumetric}
Brian Curless and Marc Levoy.
\newblock A volumetric method for building complex models from range images.
\newblock In \emph{SIGGRAPH}, pages 303--312, 1996.

\bibitem[Dai et~al.(2023)Dai, Zhu, Geng, Ruan, Zhang, and Wang]{dai2023graspnerf}
Qiyu Dai, Yan Zhu, Yiran Geng, Ciyu Ruan, Jiazhao Zhang, and He Wang.
\newblock {GraspNeRF}: multiview-based 6-dof grasp detection for transparent and specular objects using generalizable nerf.
\newblock In \emph{ICRA}, pages 1757--1763, 2023.

\bibitem[Fang et~al.(2021)Fang, Xie, Wang, Zhang, Liu, and Tian]{fang2021neusample}
Jiemin Fang, Lingxi Xie, Xinggang Wang, Xiaopeng Zhang, Wenyu Liu, and Qi Tian.
\newblock Neusample: Neural sample field for efficient view synthesis.
\newblock \emph{arXiv}, 2021.

\bibitem[Heo et~al.(2023)Heo, Kim, Lee, Lee, Kim, Kim, and Kim]{Heo2023robust}
Hwan Heo, Taekyung Kim, Jiyoung Lee, Jaewon Lee, Soohyun Kim, Hyunwoo~J. Kim, and Jin-Hwa Kim.
\newblock Robust camera pose refinement for multi-resolution hash encoding.
\newblock \emph{ICML}, 2023.

\bibitem[Jeong et~al.(2021)Jeong, Ahn, Choy, Anandkumar, Cho, and Park]{jeong2021self}
Yoonwoo Jeong, Seokjun Ahn, Christopher Choy, Anima Anandkumar, Minsu Cho, and Jaesik Park.
\newblock Self-calibrating neural radiance fields.
\newblock In \emph{ICCV}, pages 5846--5854, 2021.

\bibitem[Li et~al.(2023)Li, Gao, Tancik, and Kanazawa]{li2023nerfacc}
Ruilong Li, Hang Gao, Matthew Tancik, and Angjoo Kanazawa.
\newblock Nerfacc: Efficient sampling accelerates nerfs.
\newblock \emph{arXiv}, 2023.

\bibitem[Lin et~al.(2021)Lin, Ma, Torralba, and Lucey]{lin2021barf}
Chen-Hsuan Lin, Wei-Chiu Ma, Antonio Torralba, and Simon Lucey.
\newblock Barf: Bundle-adjusting neural radiance fields.
\newblock \emph{ICCV}, 2021.

\bibitem[Lin et~al.(2023)Lin, M{\"u}ller, Tremblay, Wen, Tyree, Evans, Vela, and Birchfield]{lin2023pnerf}
Yunzhi Lin, Thomas M{\"u}ller, Jonathan Tremblay, Bowen Wen, Stephen Tyree, Alex Evans, Patricio~A. Vela, and Stan Birchfield.
\newblock Parallel inversion of neural radiance fields for robust pose estimation.
\newblock In \emph{ICRA}, 2023.

\bibitem[Maggio et~al.(2023)Maggio, Abate, Shi, Mario, and Carlone]{maggio2023loc}
Dominic Maggio, Marcus Abate, Jingnan Shi, Courtney Mario, and Luca Carlone.
\newblock {Loc-NeRF}: {Monte Carlo} localization using neural radiance fields.
\newblock In \emph{ICRA}, pages 4018--4025, 2023.

\bibitem[Meng et~al.(2021)Meng, Chen, Luo, Wu, Su, Xu, He, and Yu]{meng2021gnerf}
Quan Meng, Anpei Chen, Haimin Luo, Minye Wu, Hao Su, Lan Xu, Xuming He, and Jingyi Yu.
\newblock Gnerf: Gan-based neural radiance field without posed camera.
\newblock In \emph{ICCV}, pages 6351--6361, 2021.

\bibitem[Mildenhall et~al.(2019)Mildenhall, Srinivasan, Ortiz-Cayon, Kalantari, Ramamoorthi, Ng, and Kar]{mildenhall2019local}
Ben Mildenhall, Pratul~P Srinivasan, Rodrigo Ortiz-Cayon, Nima~Khademi Kalantari, Ravi Ramamoorthi, Ren Ng, and Abhishek Kar.
\newblock Local light field fusion: Practical view synthesis with prescriptive sampling guidelines.
\newblock \emph{ACM TOG}, 38\penalty0 (4):\penalty0 1--14, 2019.

\bibitem[Mildenhall et~al.(2020)Mildenhall, Srinivasan, Tancik, Barron, Ramamoorthi, and Ng]{mildenhall2020nerf}
Ben Mildenhall, Pratul~P. Srinivasan, Matthew Tancik, Jonathan~T. Barron, Ravi Ramamoorthi, and Ren Ng.
\newblock Nerf: Representing scenes as neural radiance fields for view synthesis.
\newblock In \emph{ECCV}, 2020.

\bibitem[M\"uller(2021)]{tinycudann}
Thomas M\"uller.
\newblock {tiny-cuda-nn}, 2021.

\bibitem[M\"uller et~al.(2022)M\"uller, Evans, Schied, and Keller]{mueller2022instant}
Thomas M\"uller, Alex Evans, Christoph Schied, and Alexander Keller.
\newblock Instant neural graphics primitives with a multiresolution hash encoding.
\newblock \emph{ACM TOG}, 41\penalty0 (4):\penalty0 102:1--102:15, 2022.

\bibitem[Piala and Clark(2021)]{piala2021terminerf}
Martin Piala and Ronald Clark.
\newblock Terminerf: Ray termination prediction for efficient neural rendering.
\newblock In \emph{3DV}, pages 1106--1114, 2021.

\bibitem[Sch\"{o}nberger and Frahm(2016)]{schoenberger2016sfm}
Johannes~Lutz Sch\"{o}nberger and Jan-Michael Frahm.
\newblock Structure-from-motion revisited.
\newblock \emph{CVPR}, 2016.

\bibitem[Sch\"{o}nberger et~al.(2016)Sch\"{o}nberger, Zheng, Pollefeys, and Frahm]{schoenberger2016mvs}
Johannes~Lutz Sch\"{o}nberger, Enliang Zheng, Marc Pollefeys, and Jan-Michael Frahm.
\newblock Pixelwise view selection for unstructured multi-view stereo.
\newblock \emph{ECCV}, 2016.

\bibitem[Sucar et~al.(2021)Sucar, Liu, Ortiz, and Davison]{sucar21imap}
Edgar Sucar, Shikun Liu, Joseph Ortiz, and Andrew Davison.
\newblock {iMAP}: Implicit mapping and positioning in real-time.
\newblock In \emph{ICCV}, 2021.

\bibitem[Sun et~al.(2022)Sun, Sun, and Chen]{sun2022direct}
Cheng Sun, Min Sun, and Hwann-Tzong Chen.
\newblock Direct voxel grid optimization: Super-fast convergence for radiance fields reconstruction.
\newblock In \emph{CVPR}, pages 5459--5469, 2022.

\bibitem[Tang et~al.(2023)Tang, Sundaralingam, Tremblay, Wen, Yuan, Tyree, Loop, Schwing, and Birchfield]{tang2023rgb}
Zhenggang Tang, Balakumar Sundaralingam, Jonathan Tremblay, Bowen Wen, Ye Yuan, Stephen Tyree, Charles Loop, Alexander Schwing, and Stan Birchfield.
\newblock {RGB}-only reconstruction of tabletop scenes for collision-free manipulator control.
\newblock In \emph{ICRA}, pages 1778--1785, 2023.

\bibitem[Teschner et~al.(2003)Teschner, Heidelberger, Müller, Pomeranets, and Gross]{Teschner2003spatialhash}
Matthias Teschner, Bruno Heidelberger, Matthias Müller, Danat Pomeranets, and Markus Gross.
\newblock Optimized spatial hashing for collision detection of deformable objects.
\newblock \emph{In Proceedings of VMV’03, Munich, Germany}, 2003.

\bibitem[Wang et~al.(2021)Wang, Wu, Xie, Chen, and Prisacariu]{wang2021nerf--}
Zirui Wang, Shangzhe Wu, Weidi Xie, Min Chen, and Victor~Adrian Prisacariu.
\newblock Nerf--: Neural radiance fields without known camera parameters.
\newblock \emph{arXiv}, 2021.

\bibitem[Wu et~al.(2022)Wu, Lee, Bhattad, Wang, and Forsyth]{wu2022diver}
Liwen Wu, Jae~Yong Lee, Anand Bhattad, Yu-Xiong Wang, and David Forsyth.
\newblock Diver: Real-time and accurate neural radiance fields with deterministic integration for volume rendering.
\newblock In \emph{CVPR}, pages 16200--16209, 2022.

\bibitem[Yen-Chen et~al.(2020)Yen-Chen, Florence, Barron, Rodriguez, Isola, and Lin]{yen2020inerf}
Lin Yen-Chen, Pete Florence, Jonathan~T. Barron, Alberto Rodriguez, Phillip Isola, and Tsung-Yi Lin.
\newblock {iNeRF}: Inverting neural radiance fields for pose estimation.
\newblock \emph{arxiv}, 2020.

\bibitem[Yu et~al.(2022)Yu, Fridovich-Keil, Tancik, Chen, Recht, and Kanazawa]{yu2021plenoxels}
Alex Yu, Sara Fridovich-Keil, Matthew Tancik, Qinhong Chen, Benjamin Recht, and Angjoo Kanazawa.
\newblock Plenoxels: Radiance fields without neural networks.
\newblock \emph{CVPR}, 2022.

\bibitem[Zhang et~al.(2020)Zhang, Riegler, Snavely, and Koltun]{zhang2020nerf++}
Kai Zhang, Gernot Riegler, Noah Snavely, and Vladlen Koltun.
\newblock Nerf++: Analyzing and improving neural radiance fields.
\newblock \emph{CoRR}, abs/2010.07492, 2020.

\bibitem[Zhang et~al.(2018)Zhang, Isola, Efros, Shechtman, and Wang]{zhang2018perceptual}
Richard Zhang, Phillip Isola, Alexei~A Efros, Eli Shechtman, and Oliver Wang.
\newblock The unreasonable effectiveness of deep features as a perceptual metric.
\newblock \emph{CVPR}, 2018.

\bibitem[Zhu et~al.(2022)Zhu, Peng, Larsson, Xu, Bao, Cui, Oswald, and Pollefeys]{Zhu22nice}
Zihan Zhu, Songyou Peng, Viktor Larsson, Weiwei Xu, Hujun Bao, Zhaopeng Cui, Martin~R. Oswald, and Marc Pollefeys.
\newblock Nice-slam: Neural implicit scalable encoding for slam.
\newblock In \emph{CVPR}, 2022.

\bibitem[Zhu et~al.(2023)Zhu, Chen, Wu, Hou, Shi, Li, Li, Zhao, and Zhou]{zhu2023latitude}
Zhenxin Zhu, Yuantao Chen, Zirui Wu, Chao Hou, Yongliang Shi, Chuxuan Li, Pengfei Li, Hao Zhao, and Guyue Zhou.
\newblock {LATITUDE}: Robotic global localization with truncated dynamic low-pass filter in city-scale nerf.
\newblock In \emph{ICRA}, pages 8326--8332, 2023.

\end{thebibliography}
